\documentclass[a4paper]{svproc}
\usepackage{times}

\usepackage[bookmarks=true]{hyperref}

\usepackage{graphicx,subfigure}
\usepackage[font=small]{caption}
\usepackage{amssymb}
\usepackage{amsmath}
\usepackage{comment}
\usepackage{enumerate}
\usepackage{listings}
\usepackage{url}
\usepackage{epstopdf}
\usepackage{array}
\usepackage{amsmath}
\usepackage{algorithm}
\usepackage{varwidth}
\usepackage{algpseudocode}
\usepackage{multirow}
\usepackage{flexisym}
\usepackage[mathscr]{euscript}

\usepackage[utf8]{inputenc}
\usepackage{graphicx}
\usepackage{xcolor}

\usepackage{algorithm}
\usepackage{subfigure}
\usepackage{algpseudocode}
\usepackage{graphicx,subfigure}
\usepackage{hyphenat}

\graphicspath{{./figs/}}

\newcommand*\samethanks[1][\value{footnote}]{\footnotemark[#1]}
\newcommand{\cmap}{C-map}
\newcommand{\ours}{c2g-HOF}
\newcommand{\ctog}{cost-to-go}

\date{April 2020}

\begin{document}
\mainmatter              
\title{Learning to Generate Cost-to-Go Functions \\for Efficient Motion Planning \\ 
\normalsize{[Extended Abstract]}
}


\titlerunning{\ours}  

\author{Jinwook Huh\thanks{The first two authors contributed equally to this work.  saic-ny@samsung.com}, Galen Xing\samethanks, 
Ziyun Wang, Volkan Isler, and Daniel D. Lee}
\authorrunning{Huh et al.} 
\institute{Samsung AI Center, New York, NY, 10111, USA
}
\maketitle     
\vspace{-5mm}
\begin{abstract}
Traditional motion planning is computationally burdensome for practical robots, involving extensive collision checking and considerable iterative propagation of cost values.  We present a novel neural network architecture which can directly generate the \ctog{} (c2g) function for a given configuration space and a goal configuration. The output of the network is a continuous function whose gradient in configuration space can be directly used to generate trajectories in motion planning without the need for protracted iterations or extensive collision checking. This higher order function (i.e. a function generating another function) representation lies at the core of our motion planning architecture, \textbf{\ours{}}, which can take a workspace as input, and generate the \ctog{} function over the configuration space map (\cmap). Simulation results for 2D and 3D environments show that \ours{} can be orders of magnitude faster at execution time than methods which explore the configuration space during execution. We also present an implementation of \ours{} which generates trajectories for robot manipulators directly from an overhead image of the workspace. 

\end{abstract}
\section{Motivation, Problem Definition, and Related Work}
\vspace{-3mm}
Motion planning is the task of generating robot commands which take a robot from an initial configuration to a goal configuration while avoiding collisions. It is one of the fundamental problems in robotics and has been extensively studied~\cite{latombe1991robot,choset2005principles,lavalle2006planning}. Motion planning algorithms reason about the the physical \emph{workspace} the robot operates in as well as the robot's \emph{configuration space} which relate the robot pose to the physical space. Workspace and configuration space information can be combined into a free configuration space map (\cmap) which partitions the space into regions that either are in collision or are free~\cite{lavalle2006planning}.

For low degree-of-freedom (dof) robots, it is possible to build a \cmap{} explicitly as a grid and use shortest path algorithms such as Dijkstra's or A* to plan for robot motion.
Approaches such as search-based planning methods seek to plan motions directly from sensor input~\cite{bonet2001planning,maximARA}. 
Potential field methods focus on designing local control laws based upon heuristic functions which drive the robot to the goal while steering away from the obstacles~\cite{khatib1986real,rimon1992exact}. Sampling based methods use configuration space samples to build a graph-based representation of the \cmap, while RRTs explore the C-space by sampling robot inputs and simulating robot motion~\cite{lavalle2006planning,kavraki1996probabilistic,karaman2011anytime}.

\setlength\belowcaptionskip{-3mm}
\begin{figure}[!ht]
\centering
\includegraphics[width=\textwidth]{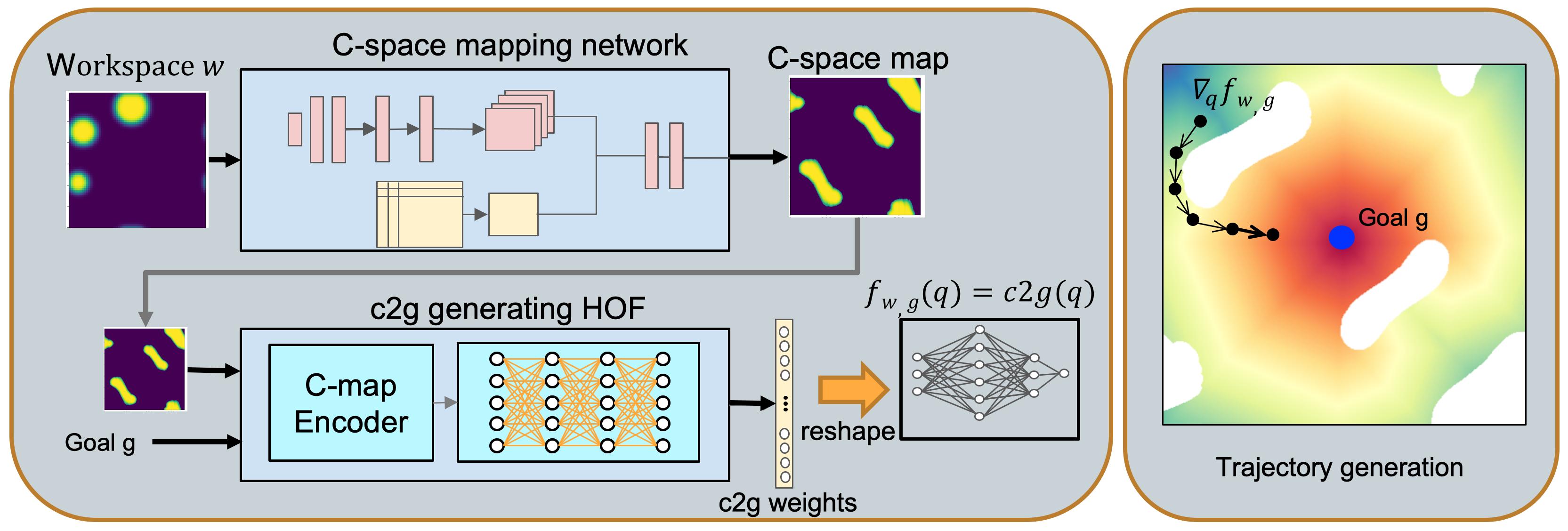}
\caption{ {\bf Left:} \ours{} architecture. The input is a representation (e.g. an image) of the workspace. The C-space mapping network builds the configuration space map. This \cmap{} along with a goal configuration are the inputs to the higher order function (HOF) network whose outputs are the weights of the \ctog{} function. These weights are reshaped into a neural network which maps robot configurations to their \ctog{} values in a continuous fashion. {\bf Right:}~Robot trajectories can be generated simply by following the gradient of the \ctog{} function with respect to the robot configuration.
\label{fig:total_network}}
\vspace{-4mm}
\end{figure}



In this work, we present a learning-based method for generating continuous motion plans. Specifically, we present a Higher Order Function (HOF) network~\cite {mitchell2019higher,wang2020geodesic}
which generates a continuous \ctog{} function over the entire configuration space. The key concept underlying our method is the notion of a Higher Order Function (HOF). In our specific case, we train a HOF to learn to output the \ctog{} function over the \cmap.  Both the HOF and the \ctog{} function are represented as neural networks. 

As shown in Figure~\ref{fig:total_network}, 
\ours{} starts with a representation of the workspace as input and generates the \cmap. The HOF then generates the \ctog{} function for a given goal over this \cmap. The HOF network generates the weights of the neural network representing the \ctog{} as a smooth and continuous function. We call our method \textbf{\ours{}} which stands for \emph{cost-to-go Higher Order Function Network}. 
The motion plan can be generated by following the gradient of the output \ctog{} function (Figure~\ref{fig:total_network}-right). 
The main advantage of our method is that, once trained, it can generate a continuous \ctog{} map over the entire \cmap{} much faster than approaches which construct \cmap{} representations during execution. 
We demonstrate the approach using a novel experimental setup where the workspace is directly represented from an overhead image in which a custom 3-dof manipulator operates. Results from synthetic experiments demonstrate that our method is more than 14 times faster than RRT and 5 times faster than A* for trajectory generation. On average, our generated trajectory lengths are only 5\% longer than the optimal A* path.


Recently, several learning-based approaches for motion planning based on deep neural networks have been proposed \cite{ichter2018learning,qureshi2018motion,molinalearning}. 
The approach of \cite{ichter2018learning} is based on a Conditional Variational Autoencoder (CVAE) for sampling policies whereas \cite{qureshi2018motion} learns motion planning networks by encoding the workspace into a latent feature space and generating actions. Our method distinguishes itself from these methods as it can estimate \ctog{} values over the entire C-space from the workspace without heavy computation. Collision free trajectories can be generated  without mapping between the workspace and the C-space using inverse and forward kinematics during execution. 
Another method is~\cite{tamar2016value} which encodes the dynamic programming iterations for motion planning in the layers of a neural network. In contrast, \ours{} learns to represent a nearly optimal cost-to-go function as the weights of a multilayer perceptron neural network.
In this extended abstract, we present the main ideas of our technical approach as well as results from ongoing experiments. 
\vspace{-3mm}


\section{Technical Approach}
\vspace{-3mm}

The \ours{} architecture, presented in Figure~\ref{fig:total_network} consists of three components: (1)~The C-map network which builds a representation of the \cmap{} from the workspace, (2)~HOF network which outputs the weights of the \ctog{} function, and (3)~the \ctog{} function represented as a neural network which can map any configuration to a positive real number which represents the distance of that configuration to the goal configuration (i.e., the \ctog{} value).



The C-map network is a functional representation of the \cmap{} as a function $f_c : \mathcal{W} \rightarrow \mathcal{C}$  where $\mathcal{W}$ is the workspace, and the output is a representation of the \cmap. In this work, we use a uniform representation with a (soft) collision indicator for each configuration. 


\setlength\belowcaptionskip{-5mm}
\begin{figure}[t]
\centering
\subfigure{\includegraphics[width=0.89\textwidth]{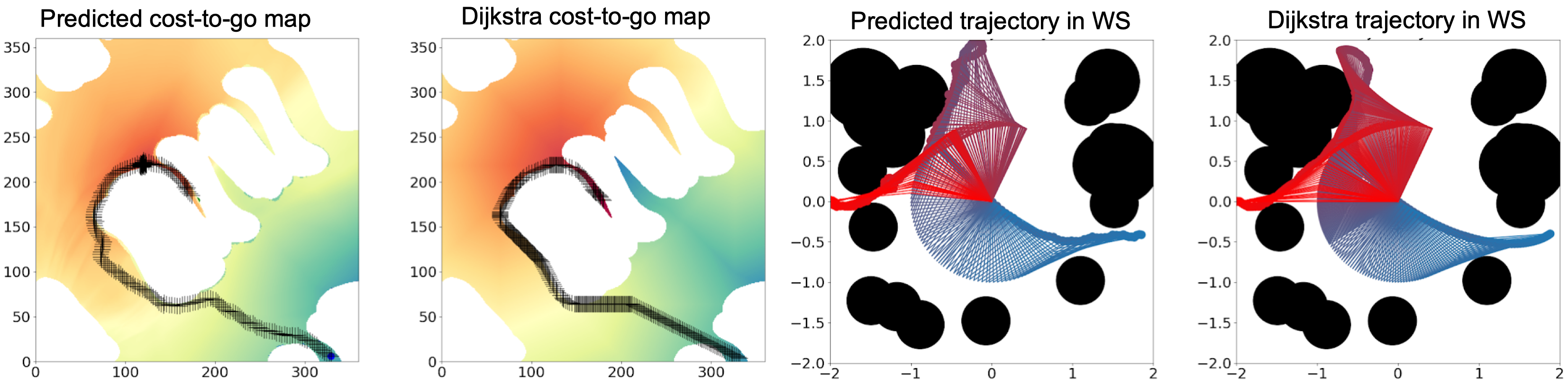} 
\label{fig:planning_2d}}\vspace*{-4mm}
\vspace{-4mm}
\subfigure{\includegraphics[width=0.88\textwidth]{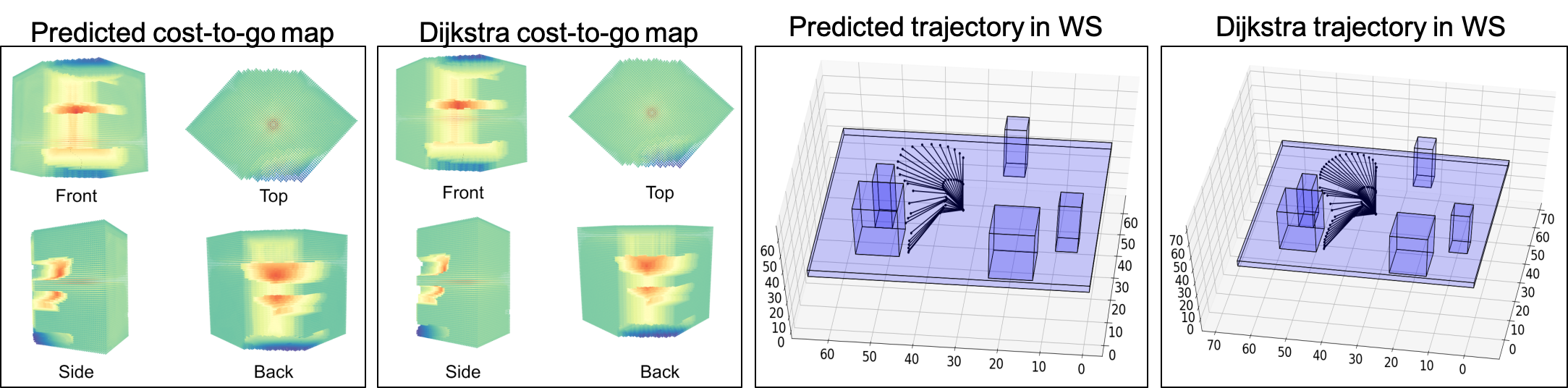}
\label{fig:planning_3d}}
\caption{Comparison of \ours{} output against the optimal solution given by Dijskstra's algorithm for 2D (top row) and 3D (bottom row) instances. 
\ctog{} map and the output trajectory are shown on the left. The same trajectory in the workspace is shown on the right.}
\label{fig:result_cost_2d}
\end{figure}

The second component is the HOF network which is trained to output the \ctog{} function from the input \cmap.
The specific encoder used in this component depends on the representation of the \cmap. In this extended abstract, we apply image and voxel based encoders for 2D and 3D representations. Other representations are possible: for example, for higher dimensions, C-space samples can be directly encoded using Point-Net or similar encoders~\cite{qi2017pointnet}. It is worth emphasizing that the output \ctog{} function is a continuous function and is not directly linked to the input representation. This is because, during training, a denser set of samples continuously sampled from the \cmap{} are used to adjust the weights of the neural network representing the \ctog{} function given in the third layer. To summarize, the HOF encoder takes the \cmap{} as input $I$ and generates a \ctog{} function $f_I : \mathcal{Q} \rightarrow \mathcal{C}$, such that  
for all $q \in \mathcal{Q}$, 
$c = f_I(q) \in \mathcal{C}$ is the \ctog{} of $q$. Motion planning can be performed by following the gradient of $f_I$. In this work, we use a three layer perceptron with 128 neurons at each layer followed by ReLU activations to represent $f_I$. 


In order to train the network, we generated a dataset composed of 30,000 randomly generated workspaces and computed \ctog{} values in C-space using Dijkstra's algorithm. Technical details on the architecture, dataset generation and training will be presented in the full version of the paper. At each iteration, given a set of 2,000 (40,000 for 3D) random samples, the network is trained by minimizing mean-square error (MSE) between the predicted \ctog{} by network and ground truth \ctog{} by Dijkstra. In the next section, we present results for two representative environments. \vspace{-4mm}






\section{Results}
\vspace{-3mm}

We evaluated \ours{} in two sets of environments. The first environment consists of a 2-link manipulator operating in a 2D random workspace and the second environment is for a 3-dof manipulator operating in a 3D workspace with randomly generated obstacles on a table. For each setting, we generated a dataset composed of 10,000 trajectories (200 workspaces $\times$ 50 random trajectories in each workspace) separately. 



\setlength\belowcaptionskip{-5mm}
\begin{figure}[t]
\centering
\subfigure{\includegraphics[width=0.31\textwidth]{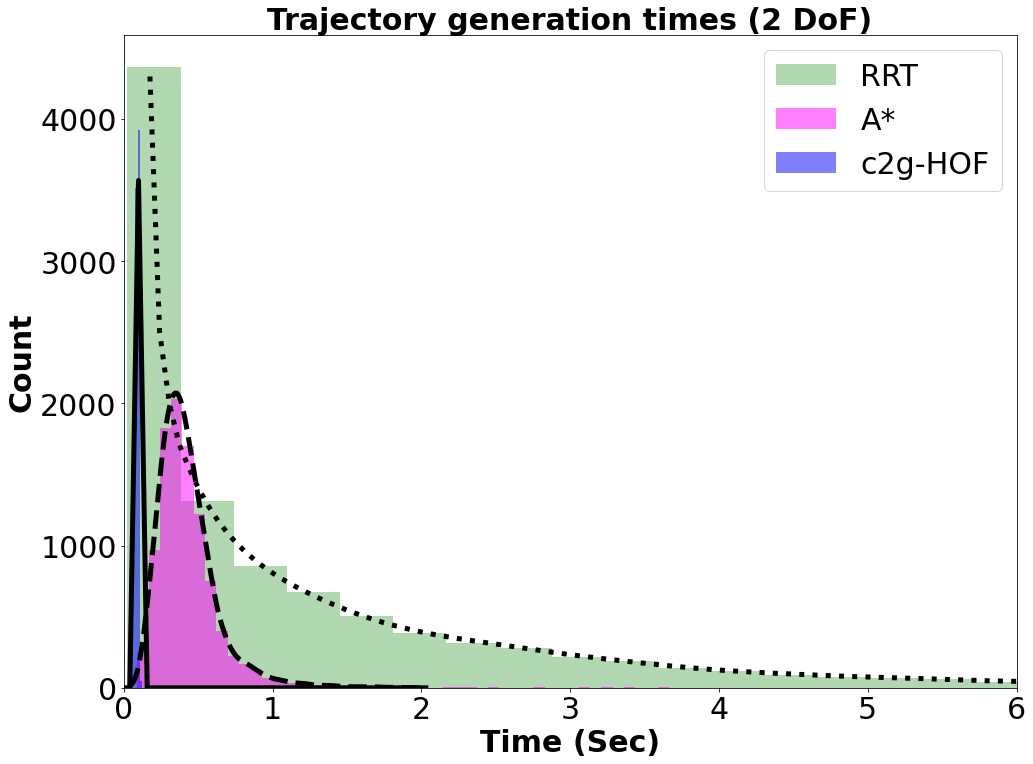} \label{fig:traj_time_histo_2d}}
\vspace{-4mm}
\subfigure{\includegraphics[width=0.31\textwidth]{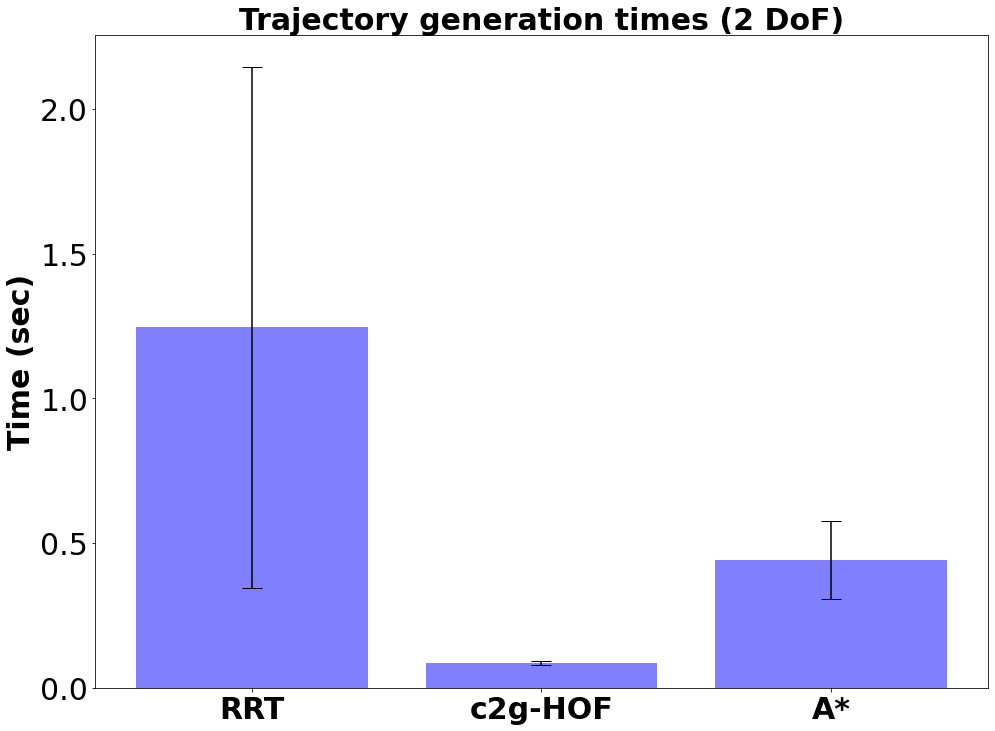} \label{fig:traj_time_2d}}
\subfigure{\includegraphics[width=0.31\textwidth]{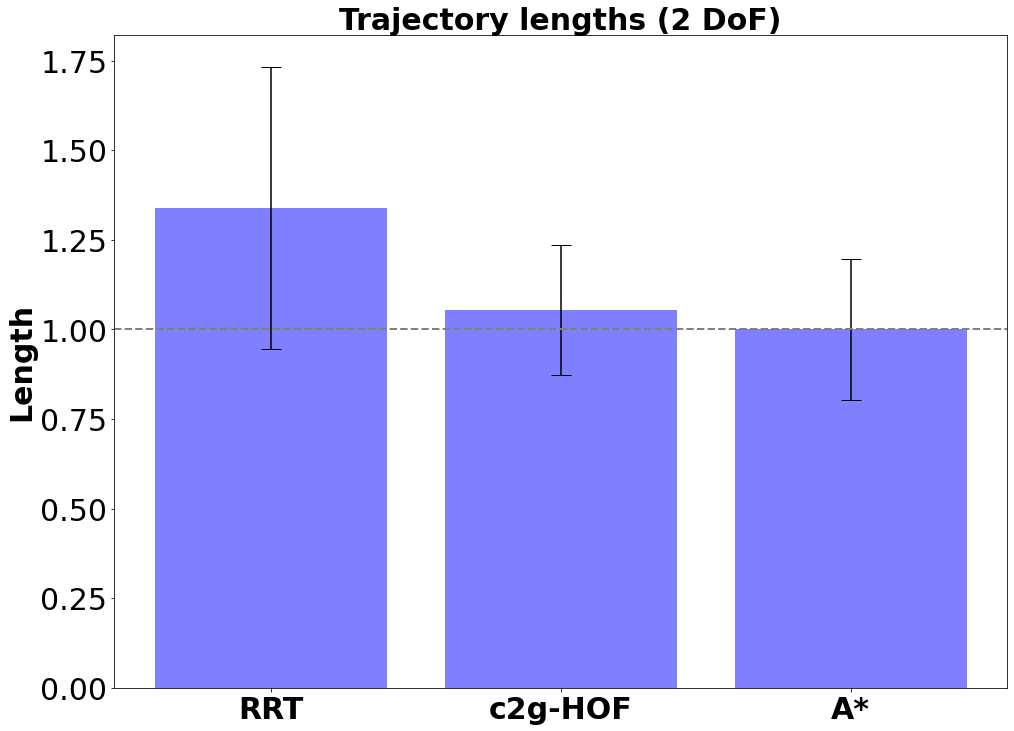}
\label{fig:traj_length_2d}}
\subfigure{\includegraphics[width=0.31\textwidth]{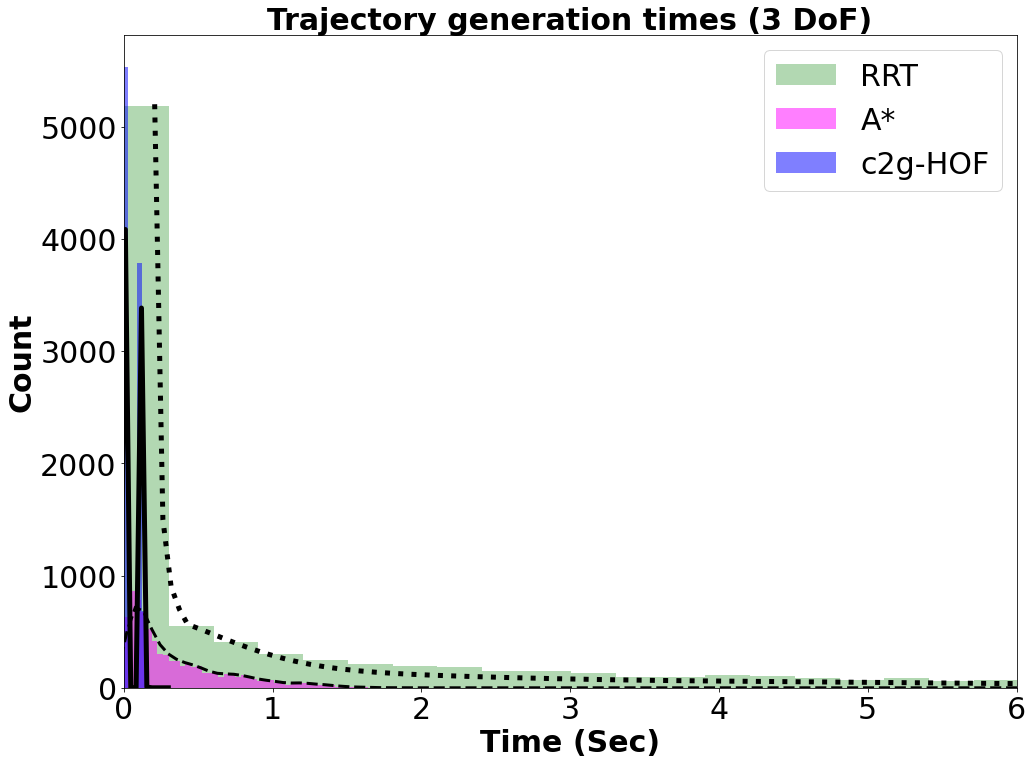} \label{fig:traj_time_histo_3d}}
\subfigure{\includegraphics[width=0.31\textwidth]{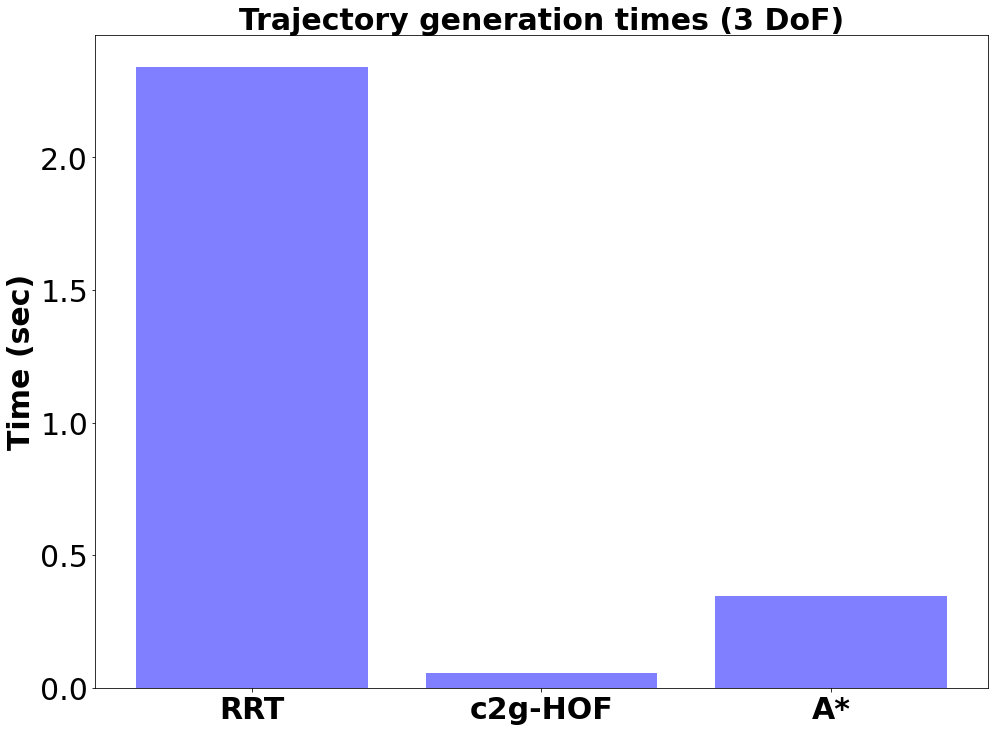}
\label{fig:traj_time_3d}}
\subfigure{\includegraphics[width=0.31\textwidth]{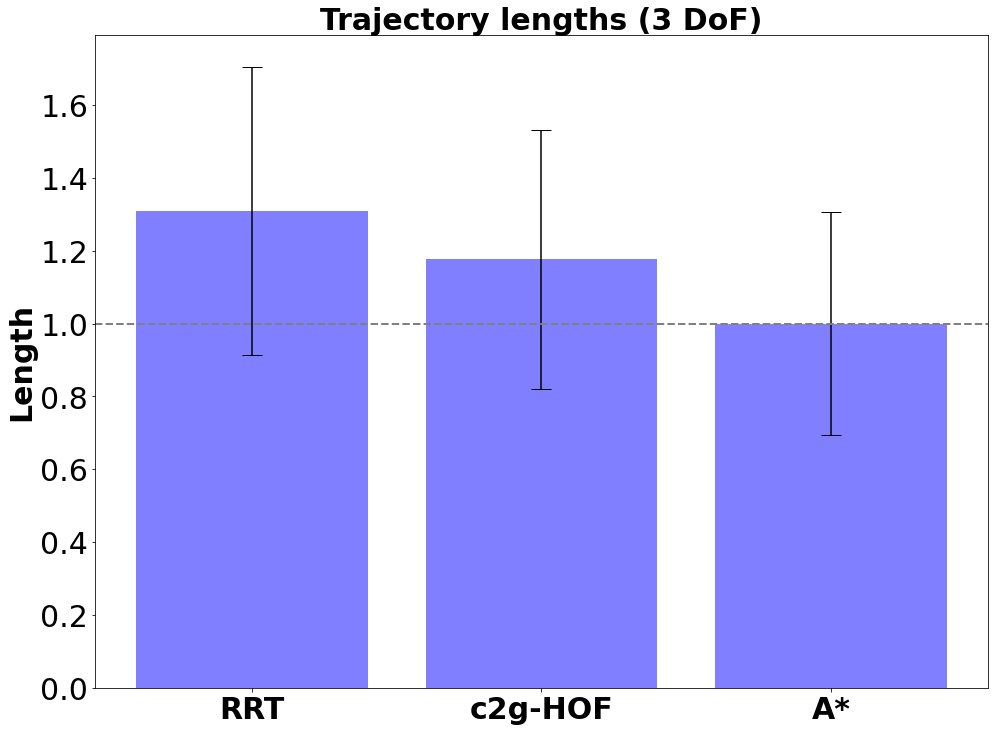}
\label{fig:traj_length_3d}}
\vspace{-4mm}
\caption{Results for the 2-dof robot (top row) and 3-dof robot (bottom row). Histogram of planning times are shown on the left. The middle plot shows mean and variances of these distributions whereas the right plot shows the trajectory length mean and variance values. (The trajectory length is normalized by the length of the A*).}
\label{fig:result_table_2d_3d}
\end{figure}


As shown in Fig.~\ref{fig:result_cost_2d}, the predicted \ctog{} value is almost the same the optimal value given by Dijkstra's algorithm. The generated trajectory arrives at the goal point while avoiding obstacles by following the gradient of \ctog{} values. Both of \ours{} and RRT have 99\% success rates.
Figure \ref{fig:result_table_2d_3d} shows comparison results between \ours{}, RRT, and A* in terms of planning time and trajectory length. While RRT and A* require many collision checks, the c2g-network computes \ctog{} without extensive collision checks or space discretization. In the 2D workspace, we see that planning of \ours{} is 14 times faster than RRT, and 5 times faster than A* with a smaller variance. The values shown are deviation from the optimal value: the trajectory length is normalized by the length of the A* in the same environments. We can see that the trajectory with \ours{} is 5\% longer than the optimal A* trajectory length but 21\% shorter than the RRT trajectory. In the 3D workspace, \ours{} is 4 times faster than A* and it generates 17\% longer trajectory. It is worth noting that the planning time of \ours{} has low variance and hence does not depend heavily on the specific instance.
\vspace{-3mm}

\section{Experiments}
\begin{figure}
\centering
\subfigure{\includegraphics[width=0.44\textwidth]{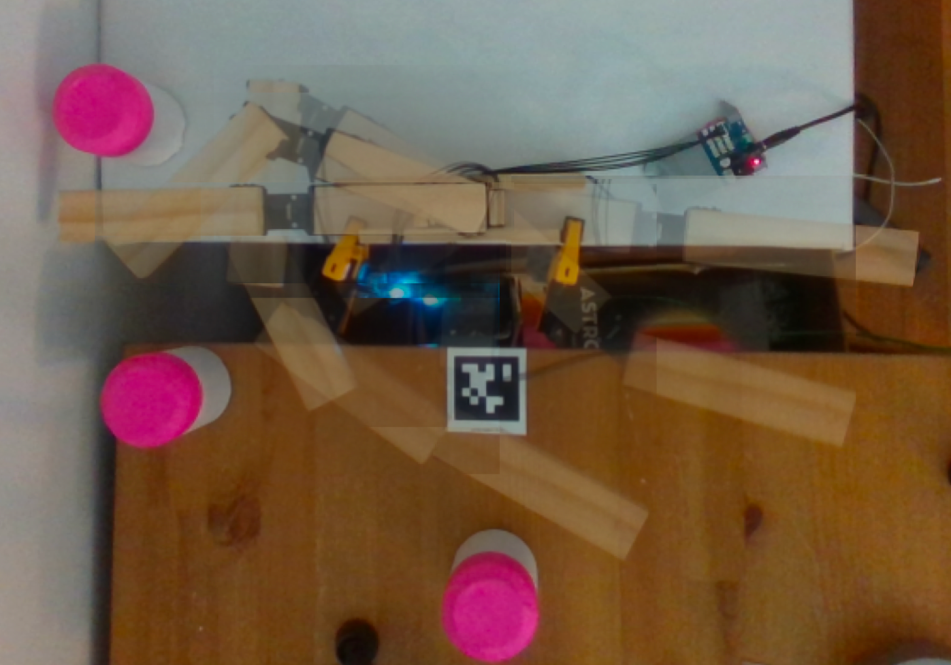} \label{fig:2d_exp}}
\subfigure{\includegraphics[width=0.42\textwidth]{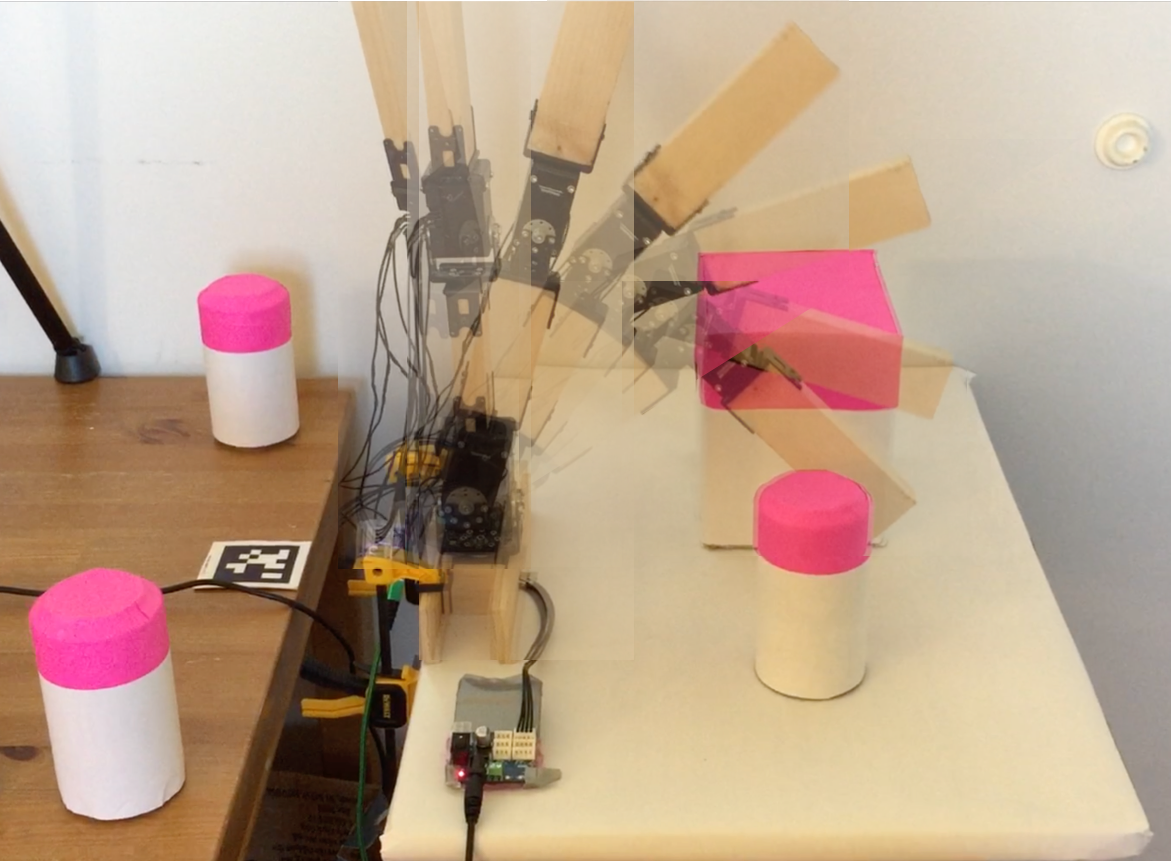}
\label{fig:3d_exp}}
\vspace{-5mm}
\caption{Our experiment setup and an  example robot motion generated by following the gradient of the cost2go network.}
\label{fig:result_exp}
\end{figure}

\vspace{-3mm}
To verify the performance of the trained \ours{} on a real robot, we conducted experiments with real 2-link and a 3-dof manipulators. In our setting, the workspace is represented directly from an overhead Realsense D435 camera. The robot shape and kinematics are learned separately for the C-map network. The HOF network generates the \ctog{} function. The trajectory is generated by following the gradient of \ctog{} values. As shown in Figure \ref{fig:result_exp}, the robot performs well even in the presence of narrow passages: 2-link and 3-dof manipulators insert the link between two obstacles while avoiding collisions successfully.

We are planning to perform additional  tests  in more complicated environments such as a kitchen shelf and will present the results in the full version of the paper. The shelf setting will generate additional challenges such as a higher number of local minima around tight spaces.
\vspace{-4mm}

\vspace{-5mm}


\section{Main Experimental Insights}
\vspace{-3mm}

We showed that a neural network can be trained to output a continuous \ctog{} function over a given workspace for a given robot. The architecture is fairly general since the same network can be adapted to various settings simply by changing the encoder architecture. In our case,  we used the same  c2g-network (except for the input dimension) in both  two and three dof cases. The performance of the HOF-network indicates that it can learn the \ctog{} function when coupled with an appropriate input representation. More broadly, our results suggest that the exploration efforts of sampling based methods such as RRTs can be shifted to training time. 

Successfully training the network to learn \ctog{} functions requires addressing two challenges: 
 The first is to prevent the network bias toward simpler environments which can happen for example when the training set obstacles are placed uniformly at random. To mitigate this issue, we made sure that our dataset has numerous ``narrow passage" instances. The details will be presented in the full version. The second challenge is regarding \cmap{} sampling. The HOF network is conditioned on a discrete map of the configuration space. However, during training, random samples over the entire c-space and their \ctog{} values are used to learn a smooth, continuous \ctog{} function. The sampling strategy during training has direct impact on performance. We leveraged techniques developed for sampling configuration spaces efficiently such as sampling densely around collision boundaries~\cite{shi2014spark,hsu1998finding}. Additional experimentation is needed to assess how these strategies will scale up to more complex environments and robots in the case of \ours{}.

In the current \ours{} architecture, we used a separately trained network to generate the \cmap{} from workspace and robot descriptions. However, our network is end to end trainable. In theory, it is possible to condition the network on a robot description (or even a picture of the robot) and learn the weights of the c-map network. This capability would allow us to incorporate online calibration/system identification into the architecture and constitutes an interesting avenue for future research.

\vspace{-3mm}
{\small
\bibliographystyle{splncs03}
\bibliography{references}
}

\end{document}